%% file: main.tex
\documentclass[11pt]{article}

\usepackage[final]{acl}

\usepackage{times}
\usepackage{latexsym}

\usepackage[T1]{fontenc}

\usepackage[utf8]{inputenc}

\usepackage{microtype}

\usepackage{inconsolata}

\usepackage{graphicx}
\usepackage{times}
\usepackage{latexsym}
\usepackage{amsmath}
\usepackage{amsfonts}
\usepackage{url}
\usepackage{graphicx}
\usepackage{booktabs}
\usepackage{xcolor}
\usepackage{rotating}
\usepackage{pifont}
\usepackage{multirow}
\usepackage{enumitem}
\usepackage[table]{xcolor}
\usepackage{subcaption}
\usepackage{tikz}
\usetikzlibrary{shapes.geometric, arrows, positioning}
\usepackage{hyperref}

\usepackage{tcolorbox}
\tcbuselibrary{listings,breakable}
\usepackage{listings}

\usepackage{pdfpages}

\newtcolorbox{promptbox}{
  colback=gray!10,
  colframe=gray!60,
  boxrule=0.3pt,
  arc=2pt,
  left=6pt,
  right=6pt,
  top=6pt,
  bottom=6pt,
  breakable
}

\definecolor{lightpurple}{RGB}{245,240,255}

\newtcolorbox{promptbox1}{
  colback=lightpurple,
  colframe=purple!50!black,
  boxrule=0.4pt,
  arc=3pt,
  left=6pt,
  right=6pt,
  top=6pt,
  bottom=6pt,
  breakable
}

\definecolor{lightgreenbg}{RGB}{235,250,240}

\newtcolorbox{statebox}{
  colback=lightgreenbg,
  colframe=green!50!black,
  boxrule=0.4pt,
  arc=3pt,
  left=6pt,
  right=6pt,
  top=6pt,
  bottom=6pt,
  breakable
}

\definecolor{lightredbg}{RGB}{255,240,240}

\newtcolorbox{judgebox}{
  colback=lightredbg,
  colframe=red!60!black,
  boxrule=0.4pt,
  arc=3pt,
  left=6pt,
  right=6pt,
  top=6pt,
  bottom=6pt,
  breakable
}

\definecolor{lightcyanbg}{RGB}{235,255,255}

\newtcolorbox{toolbox}{
  colback=lightcyanbg,
  colframe=cyan!60!black,
  boxrule=0.4pt,
  arc=3pt,
  left=6pt,
  right=6pt,
  top=6pt,
  bottom=6pt,
  breakable
}

\definecolor{bar_red}{HTML}{C65B5B}   
\definecolor{bar_teal}{HTML}{7FAEA8}

\DeclareRobustCommand{\tool}{%
  \ifmmode
    \text{tool}
  \else
    \textnormal{tool}
  \fi
}

\newcommand{\fuser}{f_{\text{user}}}
\newcommand{\ftool}{f_{\text{tool}}}
\newcommand{\fstate}{f_{\text{state}}}

\title{Toward Scalable Verifiable Reward: Proxy State-Based Evaluation for Multi-turn Tool-Calling LLM Agents}

\author{\mdseries
 Yun-Shiuan Chuang\thanks{Joint first author.}\hspace{-3pt},
 Chaitanya Kulkarni\footnotemark[1]\hspace{-3pt},
 Alec Chiu\thanks{Joint second author.}\hspace{-3pt},
\\
 Avinash Thangali\footnotemark[2]\hspace{-3pt},
 Zijie Pan\footnotemark[2]\hspace{-3pt},
 Shivani Shekhar\footnotemark[2]\hspace{-3pt},
 Yirou Ge\footnotemark[2]\hspace{-3pt},
 Yixi Li\footnotemark[2]\hspace{-3pt},
\\
 Uma Kona,
 Linsey Pang,
 Prakhar Mehrotra
\\\\
 PayPal AI
}

\begin{document}
\maketitle
\input{sections/00_abstract}
\input{sections/01_intro}

\input{sections/02_related_work}

\input{sections/03_prelim}

\input{sections/04_methods}
\input{sections/05_exp}
\input{sections/06_results}
\input{sections/07_conclusion}

\bibliography{custom}

\appendix

\input{sections/appendix}

\end{document}

%% file: sections/00_abstract.tex
\begin{abstract}
Interactive large language model (LLM) agents operating via multi-turn dialogue and multi-step tool calling are increasingly used in production. Benchmarks for these agents must both \emph{reliably compare models} and \emph{yield on-policy training data}. Prior agentic benchmarks (e.g., $\tau$-bench, $\tau^2$ bench, AppWorld) rely on fully deterministic backends, which are costly to build and iterate. We propose \emph{Proxy State-Based Evaluation}, an LLM-driven simulation framework that preserves final state-based evaluation without a deterministic database. Specifically, a \emph{scenario} specifies the user goal, user/system facts, expected final state, and expected agent behavior, and an LLM \emph{state tracker} infers a structured proxy state from the full interaction trace. LLM \emph{judges} then verify goal completion and detect tool/user hallucinations against scenario constraints. Empirically, our benchmark produces stable, model-differentiating rankings across families and inference-time reasoning efforts, and its on-/off-policy rollouts provide supervision that transfers to \emph{unseen} scenarios. Careful scenario specification yields near-zero simulator hallucination rates as supported by ablation studies. The framework also supports sensitivity analyses over user personas. Human–LLM judge agreement exceeds 90\%, indicating reliable automated evaluation. Overall, proxy state-based evaluation offers a practical, scalable alternative to deterministic agentic benchmarks for industrial LLM agents.
\end{abstract}

%% file: sections/01_intro.tex
\section{Introduction}
\label{sec:intro}

LLM-based agents are increasingly deployed to solve \emph{multi-turn}, \emph{multi-step}, \emph{tool-calling} tasks in industrial workflows (e.g., commerce, account management, customer operations). Building these agents requires two ingredients: (1) \textbf{stable evaluation} that reflects whether the agent truly accomplished user goals, and (2) \textbf{on-policy data generation} so the agent can learn from interaction, exploration, and environment feedback. Recent agentic benchmarks embrace a benchmark-as-environment paradigm (e.g., $\tau$-bench, $\tau^2$ bench, and AppWorld) \cite{yao2025taubench,barres2025tau2benchevaluatingconversationalagents,trivedi-etal-2024-appworld} with multi-turn user$\leftrightarrow$agent dialogue, multi-step agent$\leftrightarrow$tool interaction over deterministic backends, and \emph{final state-based evaluation} that scores terminal database state and user-facing responses rather than trajectory matching. While state-based evaluation accommodates the fact that there can exist multiple correct tool-calling paths, it relies on fully deterministic backends, which demands substantial engineering (schema design, deterministic tools, assertions), slowing iteration; for example, AppWorld reports $\sim$60K LOC for the engine and $\sim$40K LOC for the benchmark, plus $\sim$1.8K unit tests over 14 months \cite{trivedi-etal-2024-appworld}.

To this end, we ask: \emph{Can we retain the benefits of final state-based evaluation without building a heavy deterministic backend?} We answer with \textbf{Proxy State-Based Evaluation}, which judges success against an \emph{LLM-inferred proxy final state} extracted from the complete interaction trace (conversation + tool calls/outputs). Our use of \emph{verifiable} refers to structured, auditable evaluation grounded in explicit scenario constraints and proxy state comparisons rather than deterministic ground-truth verification. While the evaluation is LLM-mediated, it is made reliable by scenario-level specifications and validated by $>$90\% human--LLM judge agreement. Concretely, we introduce: (i) a \textit{scenario} object encoding the user goal, user/system facts, the expected final \emph{proxy} backend state, and the expected final agent reply; (ii) an \textit{LLM proxy state tracker} that infers state transitions and the proxy final state from multi-turn, multi-step traces; and (iii) an \textit{LLM judge} that compares the proxy final state and agent responses against the scenario specification to decide goal completion. In production settings, this proxy, benchmark-as-environment design can be stood up quickly and evolve with product roadmaps. It generates multi-turn, multi-step rollouts suitable for model training, provides stable, model-differentiating metrics that guide iteration, and also specifies tool schemas that inform how tools should behave even while they are still under development.

Our benchmark yields \emph{consistent capability ordering} across model families: goal completion (GC) scales with model strength and with inference-time reasoning effort. Training the RA (SFT, RFT) within the environment improves open-weight RAs using both on-policy and off-policy data. Ablation studies confirm the robustness of the proxy state tracker and to scenario completeness, and user persona variability is captured while keeping user-induced error low. Human–LLM judge agreement rate exceeds 90\%, and the  user and tool simulator hallucination rates are close to zero, supporting reliable evaluation.

\textbf{Contributions.} (1) We formalize \emph{Proxy State-Based Evaluation} and instantiate a practical benchmark that preserves state-based evaluation without a deterministic backend. (2) We propose a scenario schema and five cooperating LLM components (reasoning agent, user simulator, tool simulators, state tracker, judge). (3) We define reliability criteria and diagnostics (bootstrapped SE, hallucination rates, human–judge agreement) and execute targeted ablations (persona sensitivity; system/user fact ablations; state-tracker strength). (4) We demonstrate how the environmeny yields scalable on-policy data and off-policy data for post-training (SFT/RFT) and supports an interactive evaluation for model comparison.

\input{figures/diagram_benchmark}

%% file: figures/diagram_benchmark.tex
\begin{figure*}[tb!] 
\centering
\includegraphics[width=\linewidth]{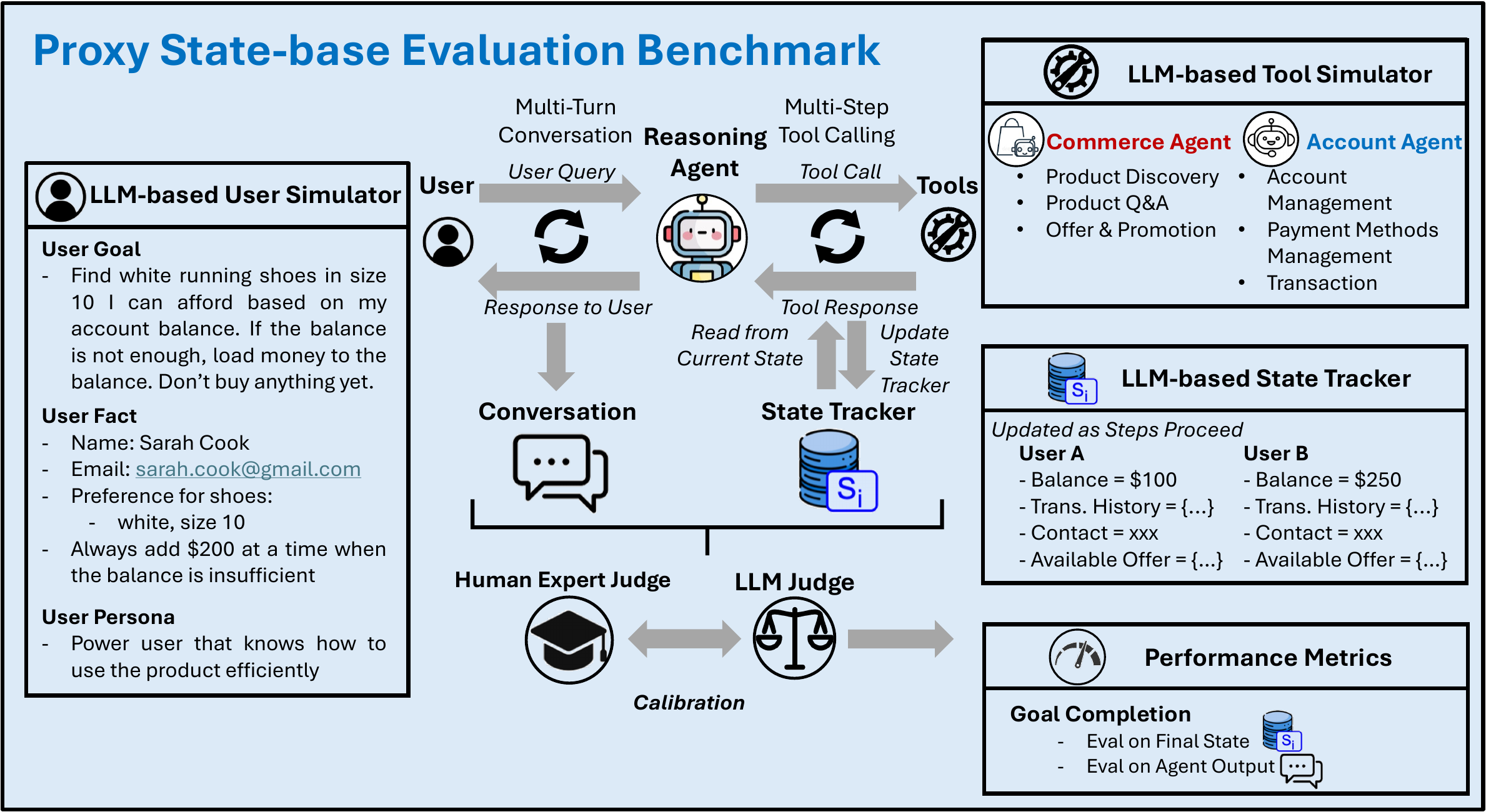}
\caption{Overview of the proxy state-based evaluation benchmark. In a \emph{multi-turn} interaction, an LLM-based user simulator converses with a reasoning agent that plans and executes \emph{multi-step} tool calls to LLM-based tool simulators. An LLM judge, calibrated with human experts, determines goal completion by checking the final proxy state. The benchmark 1) evaluates the reasoning agent’s ability to achieve goals via multi-turn dialogue and tool-calling, and also 2) yields conversation data with rewards  and supporting a leaderboard for comparing reasoning agents.}

\label{fig:exp_diagram}
\end{figure*}

%% file: sections/02_related_work.tex
\section{Related Work}
\label{sec:related}
\paragraph{State-based evaluation of interactive agents.}
Benchmarks such as $\tau$ bench, $\tau^2$ bench, and AppWorld advance \emph{final state-based evaluation}, checking terminal database state and final user response rather than trajectory matching because multiple distinct tool-calling  paths call all correctly satisfy the same user 
goal. \cite{barres2025tau2benchevaluatingconversationalagents, yao2025taubench, trivedi-etal-2024-appworld}. Our work keeps this principle but replaces the heavy deterministic backend with an LLM-inferred \emph{proxy} state and an \emph{LLM judge}. Related work in LLM-based dialogue state tracking infers \emph{dialog state} (e.g., user intents) from the conversation history \cite{carranza2025nldst, hu2022context}; however, it does not infer the \emph{backend database} state that tools read/write and that is required for state-based evaluation. In contrast, we infer a verifiable proxy database state.

\paragraph{LLMs as simulators and judges.}
LLMs have been used as user simulators, environment simulators, and automatic judges for open-ended tasks \cite{wang-etal-2024-language, zheng2023judging}. We extends this idea to \emph{state-based evaluation} by: (1) extracting a structured \textit{proxy state} from the full trace and (2) checking outcome conditions against a scenario specification.

\paragraph{On-policy data generation from simulation.}
Interactive environments support both on-policy and off-policy data generation for model training (e.g., RFT, DPO, GRPO, expert iteration) \cite{trivedi-etal-2024-appworld, chen2025retaining}. Our benchmark is designed to produce \emph{rewarded, on-policy} traces centered on tool calling and end-state verification.

%% file: sections/03_prelim.tex
\section{Preliminaries}
\label{sec:prelim}

\subsection{Task Overview}

We study the task where a \emph{reasoning agent} (RA) must achieve a user goal via \emph{multi-turn} dialogue and \emph{multi-step} tool calling.

\paragraph{Turns (user $\leftrightarrow$ RA).}
A \emph{turn} $t \in \{1,\dots,T\}$ begins with a user query $U_t$ and ends when the RA emits a user-facing text message $Y_t$ (e.g., answer, clarification, or follow-up). A task is \emph{multi-turn} if $T\!\ge\!2$.

\paragraph{Steps (RA $\leftrightarrow$ tools).}
Before the RA emits $Y_t$ to the user in turn $t$, the RA will execute $K_t$ tool steps
$S_t=((a_{t,k},r_{t,k}))_{k=1}^{K_t}$, where $a_{t,k}=\textsc{ToolCall}(\tool_{t,k},q_{t,k})$ for $\tool_{t,k}\in\mathcal{T}$ and
$r_{t,k}$ is the tool’s structured return. We say turn $t$ is \emph{multi-step}
if $K_t\ge2$. Tools in $\mathcal{T}$ take natural-language subqueries $q_{t,k}$
(LLM subagents or NL-native services, e.g., search); concrete tools are in Sec.~\ref{sec:methods}.

\paragraph{Interactive Trajectory Simulation.}
A \emph{trajectory} $\tau=(U_1,S_1,Y_1,\ldots,U_T,S_T,Y_T)$ records user utterances $U_t$, intra-turn tool-step sequences $S_t$, and agent messages $Y_t$. A \emph{user simulator} $\fuser$ generates $U_t$ conditioned on RA's message $Y_t$ and the scenario $z$; a \emph{tool simulator} $\ftool$ returns a structured output $r_{t,k}$ conditioned on the RA’s subquery $q_{t,k}$ and the scenario $z$; a \emph{state tracker} $\fstate$ updates the proxy state $\tilde{s}_{t,k}$ after each step. The scenario $z$ supplies the information these components condition on (Sec.~\ref{sec:scenario}). See Appendix~\ref{app:exp_traj} for a concrete, step-by-step example trajectory.

\paragraph{State Tracking and State-based Evaluation.}
We maintain a structured \emph{proxy state} $\tilde{s}_{t,k}$ that approximates the latent database state at step $(t,k)$. It is “proxy” because it is \emph{inferred} by an LLM state tracker $\fstate$ from tool calls rather than read from a deterministic database. The initial proxy state is initialized by the scenario ($\tilde{s}_{1,0}=s_0(z)$). After each tool step, the state tracker computes $\tilde{s}_{t,k}=\fstate(\tau_{\le (t,k)})$, where $\tau_{\le (t,k)}$ denotes the interaction history up to step $(t,k)$.\footnote{Rather than maintaining a purely step-by-step evolving state $\tilde{s}_{t,k}=f(\tilde{s}_{t,k-1},\cdot)$, which may accumulate errors over long trajectories, we condition the state tracker on the full trajectory prefix to ensure robustness.}
The tool simulator $\ftool$ “reads from” and "writes to" the current proxy state $\tilde{s}_{t,k}$. After the entire conversation finishes, state-based evaluation uses an LLM judge $J$ to check whether the final state $(\tilde{s}_T, y_T)$ satisfies the scenario’s \textit{expected final state} $s^\ast(z)$ and \textit{expected agent behavior} $b^\ast(z)$ to determine if the goal is successfully completed (details in Sec.~\ref{sec:methods}).

%% file: sections/04_methods.tex
\section{Methods}
\label{sec:methods}

\paragraph{Scenario}
\label{sec:scenario}

\input{figures/scenario}

We follow the paradigm of recent agentic benchmarks \cite{barres2025tau2benchevaluatingconversationalagents,yao2025taubench,trivedi-etal-2024-appworld}, which avoid labeling ground-truth trajectories for each task. Such labeled trajectory ignores the fact that multiple distinct tool-calling paths may correctly satisfy the same user goal. Instead, we define a \emph{scenario} $z$ that specifies outcome-level constraints rather than trajectory-level matching.

As illustrated in Fig.~\ref{fig:exp_scenario}, a scenario $z$ provides: the \emph{user goal} $g(z)$ and \emph{user facts} $u(z)$ for the user simulator $\fuser$; the \emph{system facts} $s_0(z)$ (initial database state) for the tool simulator $\ftool$, the state tracker $\fstate$, and the \emph{expected final state} and \emph{expected agent behavior} $(s^\ast(z), b^\ast(z))$ for the LLM judge. Evaluation is therefore based on whether the proxy final state and user-facing message satisfy $(s^\ast(z), b^\ast(z))$, treating all correct paths equally.

We ensure internal consistency within each scenario. The expected final state $s^\ast(z)$ must logically follow from the user goal $g(z)$ and system facts $s_0(z)$. For example, if the goal is to add \$100 to the balance and $s_0(z)$ specifies a valid funding instrument, then $s^\ast(z)$ reflects the corresponding balance increase of \$100. We ensure that all information required by the user simulator and tool simulator is fully specified in $z$. Empirically, we ensure near-zero user and tool hallucination rates in simulation (Sec.~\ref{sec:exp-settings}). All scenarios are synthetic but designed to cover diverse and realistic workflows in e-commerce and account management. Our benchmark contains $|\mathcal{Z}|=208$ scenarios, partitioned into a training set $\mathcal{Z}_{\text{train}}$ (size = 157) and a testing set $\mathcal{Z}_{\text{test}}$ (size = 51).

\paragraph{Reasoning Agent (RA) and Subagents}

The RA is the model under evaluation and also the primary training optimization target. It follows a ReAct-style loop \cite{yao2023react}: \emph{reason} $\rightarrow$ \emph{act} (tool call or show messages to the user) $\rightarrow$ observe tool/user response $\rightarrow$ next step. 

The RA’s action space comprises three calls:
\texttt{call\_account}($q$), \texttt{call\_commerce}($q$), and \texttt{show\_to\_user}($q$),
where commerce and account capabilities are served by two LLM-powered subagents that parse the subquery $q$ and return JSON outputs, and \texttt{show\_to\_user} is a special action that emits the user-facing message $Y_t$ for turn $t$ and concludes the turn $t$.


\paragraph{User simulator ($\fuser$)}
The user simulator generates the next user utterance $U_t$ conditioned on the scenario’s 
\emph{user goal} $g(z)$, \emph{user facts} $u(z)$, the selected persona 
$p\!\in\!\{\text{power},\text{ambiguous},\text{confused}\}$ 
(see Sec.~\ref{sec:exp-settings}), and the RA’s previous user-facing messages $Y$. It emits a special \texttt{<done>} token  when it believes its user goal has been fulfilled, or when the maximun turn $T_{\max}=10$ is exhausted.

\paragraph{Tool simulators ($\ftool$)}
Tool simulators implement \texttt{call\_account} and \texttt{call\_commerce} ( Fig.~\ref{fig:exp_diagram}). 
Each tool call is generated conditioned on:
(1) the \emph{system facts} $s_0(z)$ (initial database state),  
(2) the current \emph{proxy state} $\tilde{s}_{t,k-1}$, and  
(3) the RA’s subquery $q_{t,k}$. Formally, tool outputs are produced as $ r_{t,k} = \ftool(s_0(z), \tilde{s}_{t,k-1}, q_{t,k})$, 
ensuring that tools “read from” the current proxy state. Tool simulators are constrained not to fabricate information beyond the state $(s_0(z), \tilde{s}_{t,k-1})$ and the RA's subquery content $q$.

\paragraph{Proxy State Tracker}

The state tracker $\fstate$ implements the proxy state mechanism defined in Sec.~\ref{sec:prelim}. 
At each step $(t,k)$, it infers the current proxy state
$\tilde{s}_{t,k} = \fstate(\tau_{\le (t,k)})$. Tool calls are categorized as \emph{read} or \emph{write} operations.
While read-only calls (e.g., product search or transaction lookup) do not modify state, write operations (e.g., “add \$200 to balance” or “create dispute”) modify state fields only if the corresponding tool output $r_{t,k}$ indicates success. 

\paragraph{LLM Judges}

Given the \emph{proxy final state} $\tilde{s}_T$ and the entire trajectory $\tau$ (which includes all the user-facing messages), the judge evaluates them against the scenario specification $(s^\ast(z), b^\ast(z))$. Concretely, we use two LLM judges as follows.

\paragraph{(1) Goal-completion Judge.}
The primary judge classifies $(\tilde{s}_T, \tau$ into one of three outcomes: 
(i) \emph{goal completed}, 
(ii) \emph{goal not completed due to user error}, or 
(iii) \emph{goal not completed due to agent error}. 
Formally, it produces
$J_{\text{goal}}(\tilde{s}_T, y_T, \tau, z) \;\rightarrow\; \{\, c,\, e_{\text{user}},\, e_{\text{agent}} \,\}$,
where $c\!\in\!\{0,1\}$ indicates goal completion. 

\paragraph{(2) Hallucination Detection Judge.}
A separate judge detects hallucinations and returns two binary indicators:
$J_{\text{hall}}(\tau, z) \;\rightarrow\; \{\, h_{\text{tool}},\, h_{\text{user}} \,\}$,
where $h_{\text{tool}}\in\{0,1\}$ indicates \emph{tool hallucination} and 
$h_{\text{user}}\in\{0,1\}$ indicates \emph{user hallucination}. Tool hallucination is defined as the tool simulator $\ftool$ producing information that is not supported by the scenario’s system facts $s_0(z)$ or RA's subqeury $q$. User hallucination is defined as the user simulator $\fuser$ generating information that is inconsistent with the scenario’s user facts $u(z)$, the user goal $g(z)$, and RA's response $Y$ \cite{ji2023survey}.

To validate the LLM judges, we compare the goal-completion judge and the detection judge against two independent human domain expert annotations. The human–LLM judge agreement rate exceeds 90\%  (Appendix~\ref{app:human_expert_annotation}).

\paragraph{Evaluation Metrics.}
For a set of scenarios $\mathcal{Z}$ and any binary indicator $x(z)$, we define its rate as 
$\mathrm{Rate}(x)=\frac{1}{|\mathcal{Z}|}\sum_{z\in\mathcal{Z}} x(z)$. 
We report goal completion rate $\mathrm{GC}=\mathrm{Rate}(c)$, 
user-error rate $\mathrm{ER}_{\text{user}}=\mathrm{Rate}(e_{\text{user}})$,
agent-error rate $\mathrm{ER}_{\text{agent}}=\mathrm{Rate}(e_{\text{agent}})$,
tool hallucination rate $\mathrm{HR}_{\text{tool}}=\mathrm{Rate}(h_{\text{tool}})$, 
and user hallucination rate $\mathrm{HR}_{\text{user}}=\mathrm{Rate}(h_{\text{user}})$. All reported evaluation metrics are computed on $\mathcal{Z}_{\text{test}}$. Each scenario is run for 10 independent rollouts and metrics are averaged across rollouts for robust estimation.

\paragraph{Training}
We investigate two training paradigms for the LLM underlying the reasoning agent (RA), using trajectories from $\mathcal{Z}_{\text{train}}$. Only the RA’s LLM parameters are updated; other LLMs ($\fuser$,$\ftool$, $\fstate$, $J$) remain fixed. \textbf{Training with On-policy Data.} The current RA interacts with the simulator to generate trajectories $\tau$, each scored by $J_{\text{goal}}$; we retain only $c{=}1$ rollouts and use them as supervised targets for rejection-sampling fine-tuning (RFT; \citealp{anil2025rejection}) of the RA. \textbf{Training with Off-policy Data.} We replace the RA with a stronger teacher to generate trajectories on the same $\mathcal{Z}_{\text{train}}$; again, only $c{=}1$ rollouts are retained and used for supervised fine-tuning (SFT; \citealp{ouyang2022training}) of the base RA. 

%% file: figures/scenario.tex
\begin{figure}[tb!] 
\centering
\includegraphics[width=\linewidth]{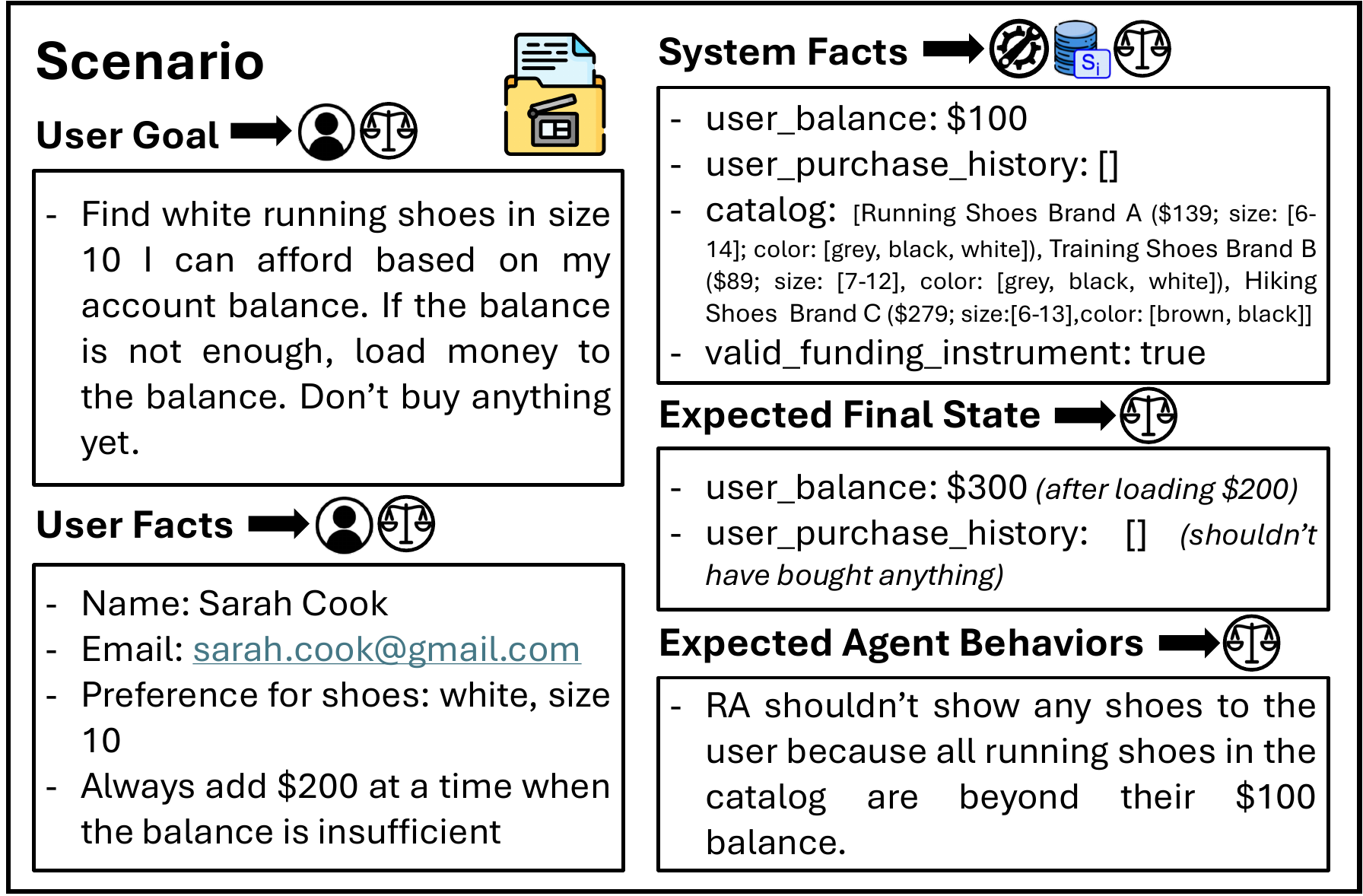}
\caption{A scenario $z$ specifies user goal $g(z)$ and user facts $u(z)$ (both used by user simulator and LLM judge), system facts $s_0(z)$ (used by tool simulators, state tracker, and LLM judge), expected final state $s^\ast(z)$, and expected agent behavior (both used by LLM judge). Arrows denote inputs. These fields drive the interactive simulation and proxy state-based evaluation.}
\label{fig:exp_scenario}

\end{figure}

%% file: sections/05_exp.tex
\section{Experimental Settings}
\label{sec:exp-settings}

\input{figures/bar_main_result}

\paragraph{Domains and Tools.}
We expose two tool families $\mathcal{T}$:
\emph{Commerce} (Product Discovery, Checkout, Cart Management, Product Q\&A, Offers \& Promotions) and
\emph{Account} (Account Management, Wallet \& Funding, Payment \& Transfer, Dispute \& Refund, Security \& Fraud, Transaction Inquiry).
The reasoning agent (RA) interacts via \texttt{call\_commerce}($q$), \texttt{call\_account}($q$), and \texttt{show\_to\_user}($q$) as defined in Sec.~\ref{sec:methods}. Each subagent governs a broad and heterogeneous set of tools.

\noindent\textbf{Reasoning Agent (RA) Models.}
We evaluate a diverse set of LLMs as the RA, including:
GPT-5 (reasoning effort $\in$ \{\emph{minimal}, \emph{low}, \emph{medium}, \emph{high}\})~\cite{singh2025openai},
GPT-4o,
GPT-4o-mini,
Gemini-2.5-pro,
Gemini-2.5-flash~\cite{comanici2025gemini},
Qwen3-235B-A22B,
and Qwen3-30B-A3B-Thinking-2507 \cite{yang2025qwen3}.
All models use temperature $=1$ during trajectory rollout.

\paragraph{Simulators and Judge Models.}
Unless otherwise specified, the user simulator $\fuser$, tool simulator $\ftool$, state tracker $\fstate$, and LLM judges $J$ are instantiated using GPT-5o with medium reasoning effort. This configuration empirically yields near-zero tool hallucination rate (1.33\%) and user hallucination rate (0.67\%).

\paragraph{User Personas.} The user simulator $\fuser$ is instantiated with a persona variable $p\!\in\!\{\text{power},\text{ambiguous},\text{confused}\}$.  Unless otherwise specified, we evaluate with $p=\text{power}$ to ensure that failure cases are primarily attributable to the RA rather than user behaviors. Detailed persona definitions are provided in Appendix~\ref{app:user_persona}.

\paragraph{Training Configuration.}
For training experiments, we use Qwen3-30B-A3B-Thinking-2507 as the base RA model and fine-tune it on $\mathcal{Z}_{\text{train}}$ as described in Sec.~\ref{sec:methods}. The primary off-policy SFT teacher is GPT-5 (reasoning effort = high), which achieves 85.76\% GC. We also experiment with Qwen3-235B-A22B as an alternative teacher (71.47\% GC) to verify that off-policy gains are not specific to GPT-5-generated data (see Sec.~\ref{sec:results}). Detailed hyperparameters and training data statistics are provided in Appendix~\ref{app:training_hyper_pars}.

%% file: figures/bar_main_result.tex
\begin{figure}[tb!] 
\centering
\includegraphics[width=\linewidth]{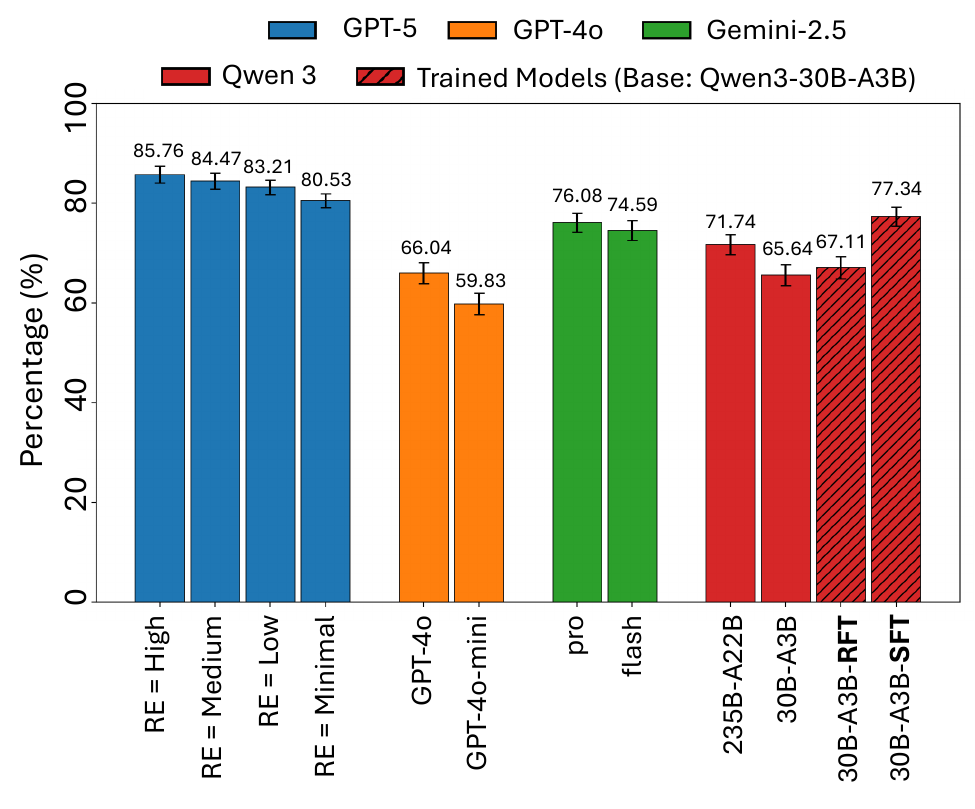}
\caption{Goal completion rate (GC) on testing scenarios $\mathcal{Z}_{\text{test}}$ across baseline reasoning agents and trained models. Error bars show the bootstrap standard error. Fine-tuning substantially improves the base Qwen3-30B-A3B-Thinking-2507 model. RE: reasoning effort.}

\label{fig:bar_main_result}
\end{figure}

%% file: sections/06_results.tex
\section{Results}
\label{sec:results}

\input{figures/exp_ablation_facts}
\input{figures/exp_user_persona}

\subsection{Goal Completion Across Models}

\paragraph{Baseline model comparison.}
Across model families (Fig.~\ref{fig:bar_main_result} ), goal completion rate (GC) scales with model strength and inference-time reasoning effort. Larger variants consistently outperform their smaller counterparts (e.g., GPT\mbox{-}4o $>$ GPT\mbox{-}4o\mbox{-}mini; Gemini-2.5\mbox{-}Pro $>$ \mbox{Flash}; Qwen3\mbox{-}235B $>$ 30B), and within the GPT-5 family, increasing reasoning effort yields a monotonic GC gain (high $>$ medium $>$ low $>$ minimal; e.g., $85.76>80.53\%$). The alignment of these trends with expected capability ordering indicates that our proxy state-based evaluation is suitable for \emph{ranking and separability} of RA's performance. 

\paragraph{Training improvements.}
We further evaluate training effects on Qwen3-30B-A3B-Thinking-2507. The base model achieves 65.64\% GC. Rejection-sampling fine-tuning (RFT) yields a modest improvement to 67.11\%, while supervised fine-tuning (SFT) using filtered successful trajectories substantially improves GC to 77.34\%. We additionally verify that off-policy gains are not specific to GPT-5-generated data: using Qwen3-235B-A22B as teacher (71.47\% GC) still improves the base model from 65.64\% to 70.87\%.

\paragraph{Multi-turn depth analysis.}
To examine how performance varies with conversation complexity, we stratify the evaluation set by turn depth (Table~\ref{tab:turn_depth}). GC decreases with depth for the base model (74.75\% for short vs.\ 42.75\% for long), confirming that deeper interactions are substantially harder. SFT improves all turn-depth buckets, with the largest absolute gain on long conversations (+26.69 pp), supporting the claim that the proxy state-based training signal generalizes to harder, multi-step interactions.

\begin{table}[t]
\centering
\small
\begin{tabular}{@{}lrcc@{}}
\toprule
\textbf{Turn depth} & \textbf{\# Scenarios (\%)} & \textbf{Base GC} & \textbf{SFT GC} \\
\midrule
Short (1--2) & 40.35 & 74.75 & 82.44 \\
Medium (3--4) & 35.04 & 69.51 & 85.39 \\
Long (5+) & 24.61 & 42.75 & 69.44 \\
\bottomrule
\end{tabular}
\caption{Goal completion rate (\%) stratified by conversation turn depth on $\mathcal{Z}_{\text{test}}$.}
\label{tab:turn_depth}
\end{table}

\subsection{Ablations}

\paragraph{Proxy State Tracker Model.}
To assess the impact of the state tracker $\fstate$, we replace the default configuration (GPT-5o, medium reasoning effort) with the weaker model GPT-4o while keeping all other components fixed. The tool hallucination rate increases from $1.33\%\pm0.53$ to $3.61\%\pm0.88$. This suggests that weaker state inference degrades consistency of the proxy state $\tilde{s}_{t,k}$, which in turn propagates errors to tool outputs since $\ftool$ reads from the current proxy state. The results highlights the importance of accurate state tracking for stable state-based evaluation. A manual validation of 50 intermediate state transitions further confirms 92\% correctness with 0.86 inter-annotator agreement (Appendix~\ref{app:human_expert_annotation}).

\paragraph{System-Fact and User-Fact Ablations.}
We further ablate scenario completeness by randomly removing a fraction of the specified \emph{system facts} $s_0(z)$ or \emph{user facts} $u(z)$ while keeping the evaluated RA fixed. Fig.~\ref{fig:exp_ablation_facts} shows a monotonic degradation: removing system facts substantially increases tool hallucination. Similarly, removing user facts increases user hallucination. These results validate that hallucination rates are sensitive to the underlying scenario specification: incomplete $s_0(z)$ induces tool-side fabrication, while incomplete $u(z)$ induces user-side fabrication. Overall, these results highlight that the meticulous effort we invest in curating scenario files is essential. It specifies the complete information required by each scenario keeps simulation grounded and minimized both tool- and user-side hallucinations.

\subsection{Cross-Model Robustness}
\label{sec:cross_model}
To test whether our findings are robust to the choice of LLMs serving multiple roles (simulator, state tracker, judge), we run two additional configurations beyond the default GPT-5o-for-all setup: (a)~simulators + state tracker = GPT-5 (medium), judge = Gemini-2.5-Pro; (b)~simulators + state tracker = Gemini-2.5-Pro, judge = Gemini-2.5-Pro. Table~\ref{tab:cross_model} reports GC across all three settings.
The main conclusions are preserved: SFT consistently improves the base RA (e.g., 65.64$\to$77.34 in the original, 53.75$\to$71.89 in Exp~(a), 64.59$\to$79.96 in Exp~(b)), and GPT-5-family models remain stronger than Gemini-2.5-family models across all settings. These results confirm that model rankings and training gains are robust to the choice of simulator and judge, ruling out circular bias from a single-family generation-evaluation setup.

\begin{table}[t]
\centering
\small
\resizebox{\columnwidth}{!}{
\begin{tabular}{@{}lccc@{}}
\toprule
\textbf{RA Model} & \textbf{Original} & \textbf{Exp (a)} & \textbf{Exp (b)} \\
\midrule
GPT-5 (high) & 85.76 & 75.78 & 86.59 \\
Gemini-2.5-Pro & 76.08 & 65.79 & 70.56 \\
Qwen3-235B-A22B & 71.74 & 62.22 & 68.61 \\
Qwen3-30B-A3B & 65.64 & 53.75 & 64.59 \\
Qwen3-30B-A3B-SFT & 77.34 & 71.89 & 79.96 \\
\bottomrule
\end{tabular}
}
\caption{Cross-model robustness: GC (\%) under different simulator/judge configurations. Original: all GPT-5o; (a): sim.\ = GPT-5 (med.), judge = Gemini-2.5-Pro; (b): all Gemini-2.5-Pro.}
\label{tab:cross_model}
\end{table}

\subsection{User Persona Sensitivity}

We evaluate the impact of user personas $p\!\in\!\{\text{power},\text{confused},\text{ambiguous}\}$ on error patterns. Fig.~\ref{fig:exp_user_persona}) reports error due to user ($\mathrm{ER}_{\text{user}}$) and user hallucination rate ($\mathrm{HR}_{\text{user}}$) across personas. When evaluated with the default \emph{power} user, the user-error rate is 3.55\% and user hallucination rate is 0.67\%. In contrast, the \emph{confused} and \emph{ambiguous} personas exhibit higher user-error rates (5.14\% and 5.50\%, respectively) and higher user hallucination rates (1.87\% and 2.75\%).  These results demonstrate that the benchmark meaningfully captures variation in user behavior. Importantly, we use power-user as our default setting (similar to \citealp{yao2025taubench}) so user-induced errors remain low. This ensures that model comparisons primarily reflect RA performance rather than user-side errors.

%% file: figures/exp_ablation_facts.tex
\begin{figure}[tb!] 
\centering
\includegraphics[width=\linewidth]{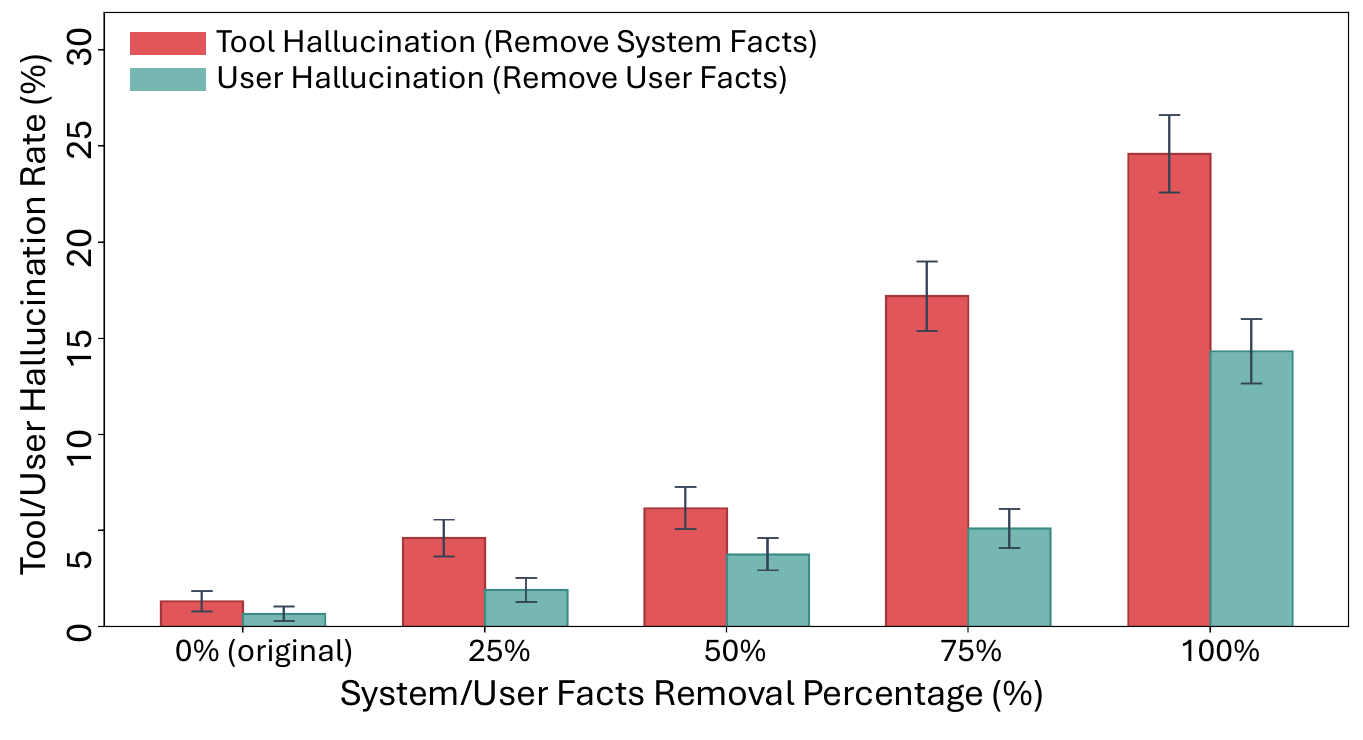}
\caption{\textbf{Ablations on scenario facts increase hallucinations.} We randomly remove a fraction of \textcolor{bar_red}{system facts} $s_0(z)$ or \textcolor{bar_teal}{user facts} $u(z)$. Tool hallucination rate and user hallucination rate rise steadily with more facts being removed. Error bars show the bootstrap standard error.}

\label{fig:exp_ablation_facts}
\end{figure}

%% file: figures/exp_user_persona.tex
\begin{figure}[tb!] 
\centering
\includegraphics[width=\linewidth]{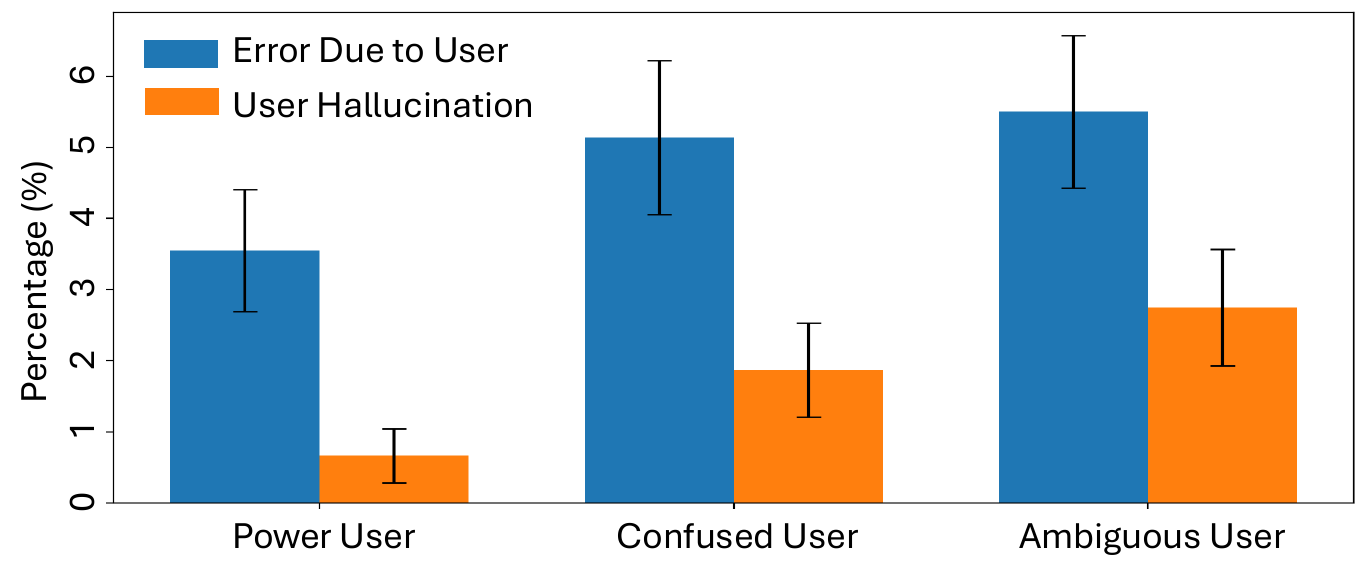}
\caption{User persona sensitivity analysis. Error due to user ($\mathrm{ER}_{\text{user}}$) and user hallucination rate ($\mathrm{HR}_{\text{user}}$) across three personas $p$. More challenging personas increase user-induced errors and user hallucination rates. Error bars denote bootstrap standard error.}

\label{fig:exp_user_persona}
\end{figure}

%% file: sections/07_conclusion.tex
\section{Conclusion}
We introduced \emph{Proxy State-Based Evaluation}, a benchmark-as-environment that preserves the benefits of final state evaluation without the engineering burden of a fully deterministic backend. Our scenario schema and cooperating LLM components (RA, $\fuser$, $\ftool$, $\fstate$, $J$) yield stable, interpretable metrics and \emph{model-differentiating rankings} across families and reasoning-effort settings. Reliability is supported by human–LLM agreement (>90\%) and near-zero simulator hallucination under the default configuration. Ablations studies show the importance of accurate state inference and scenario completeness. Beyond evaluation, the environment produces on-/off-policy rollouts that improve an open-weight RA via SFT/RFT and \emph{transfer to unseen scenarios}. Persona studies confirm meaningful sensitivity to user. Taken together, our industrial evaluation indicates that this proxy environment is a practical, scalable alternative to deterministic suites. It supports faster training iteration for LLM agents while retaining rigorous state-based evaluation.

%% file: sections/appendix.tex
\newpage
\appendix
\onecolumn

\input{sections/appendix/human_study_annotation_v2}

\input{sections/appendix/training_hyper_param}

\input{sections/appendix/user_persona}

\input{sections/appendix/conversation_flow}



%% file: sections/appendix/human_study_annotation_v2.tex
\section{Human Evaluation and Inter-Rater Agreement}
\label{app:human_expert_annotation}

\textbf{Protocol.} Two domain experts independently annotated \(n{=}50\) randomly sampled conversations from \(\mathcal{Z}_{\text{test}}\) across three dimensions: goal completion \(c\), tool hallucination \(h_{\text{tool}}\), and user hallucination \(h_{\text{user}}\).  We compare their labels to the outputs of the LLM judges \(J_{\text{goal}}\) and \(J_{\text{hall}}\) defined in the main paper.

\subsection*{Results}
Table~\ref{tab:ira_summary} reports the \emph{three-way agreement} rate, i.e., the percentage of examples where \emph{both} human annotators and the LLM judge fully agree on the label for a given dimension.

\begin{table}[h]
\centering
\begin{tabular}{lc}
\toprule
\textbf{Dimension} & \textbf{Three-way Agreement (\%)} \\
\midrule
Goal completion (\(c\)) & 82.7 \\
Tool hallucination (\(h_{\text{tool}}\)) & 94.7 \\
User hallucination (\(h_{\text{user}}\)) & 94.7 \\
\bottomrule
\end{tabular}
\caption{Three-way agreement among two human experts and the LLM judge on \(n{=}50\) conversations.}
\label{tab:ira_summary}
\end{table}

\subsection*{Takeaways}
The LLM judges align closely with human experts, supporting the reliability of our evaluation setup.

\subsection*{Intermediate State-Tracker Validation}
Beyond final-state evaluation, we conduct a separate manual study to validate the proxy state tracker on intermediate steps. We sample 50 state transitions from multi-turn conversations, each occurring immediately after a tool call. Three human annotators independently judged whether the inferred proxy state was correct. Using the human majority label as reference, the state tracker achieves 92\% correctness on these intermediate states, with 0.86 inter-annotator agreement (Fleiss' $\kappa$). This confirms that the proxy state tracker maintains high fidelity throughout the interaction, not only at conversation end.

%% file: sections/appendix/training_hyper_param.tex
\section{Training Data and Hyperparameters}
\label{app:training_hyper_pars}

\paragraph{Training data.}
We roll out each training scenario $z\!\in\!\mathcal{Z}_{\text{train}}$ ($|\mathcal{Z}_{\text{train}}|{=}157$) for 10 trajectories, yielding
$\mathcal{D}_{\text{raw}}=\{\tau_i\}_{i=1}^{1570}$.
Each trajectory $\tau$ is scored by the goal-completion judge $J_{\text{goal}}$ with indicator $c(\tau)\!\in\!\{0,1\}$.
For rejection-sampling fine-tuning (RFT) and supervised fine-tuning (SFT), we success-filter as
\[
\mathcal{D}^{\text{succ}}_{\text{rft}}=\{\tau\in\mathcal{D}_{\text{raw}}: c(\tau){=}1\},\quad
|\mathcal{D}^{\text{succ}}_{\text{rft}}|=1078;\qquad
\mathcal{D}^{\text{succ}}_{\text{sft}}=\{\tau\in\mathcal{D}_{\text{raw}}: c(\tau){=}1\},\quad
|\mathcal{D}^{\text{succ}}_{\text{sft}}|=1209.
\]
Each successful trajectory is decomposed into stepwise supervision pairs at RA emission points (tool calls or user-facing messages):
$\mathcal{S}_{\text{rft}}=\{(x_s,y_s)\}$ with $|\mathcal{S}_{\text{rft}}|=5017$, and
$\mathcal{S}_{\text{sft}}=\{(x_s,y_s)\}$ with $|\mathcal{S}_{\text{sft}}|=8057$,
where $x_s$ is the trajectory prefix immediately before emission (the history $\tau_{<s}$) and $y_s\!\in\!\{a_{t,k},Y_t\}$ is the RA’s next response.

\paragraph{Hyperparameters.}
We fine-tune the base RA model (Qwen3\mbox{-}30B\mbox{-}A3B\mbox{-}Thinking\mbox{-}2507) with LoRA~\cite{hu2022lora}:
\begin{itemize}
    \item LoRA rank/alpha: 32 / 32
    \item LoRA targets: all linear projections in self-attention (Q, K, V, O)
    \item MoE router auxiliary loss coefficient: $1{\times}10^{-3}$
    \item Learning rate: $1{\times}10^{-5}$ (constant; no scheduler)
    \item Training epochs: 2
\end{itemize}

%% file: sections/appendix/user_persona.tex
\section{User Persona Definitions}
\label{app:user_persona}

The user simulator $\fuser$ supports three persona $p$ configurations:

\begin{itemize}
    \item \textbf{Power user:} Provides complete constraints upfront and interacts efficiently.
    \item \textbf{Ambiguous user:} Initially omits key information and requires clarification.
    \item \textbf{Confused user:} Seeks guidance about the process and may misunderstand system responses.
\end{itemize}

%% file: sections/appendix/conversation_flow.tex
\section{Example Trajectory}
\label{app:exp_traj}


\begin{figure}[p]
  \centering
  \includegraphics[page=1,width=\textwidth]{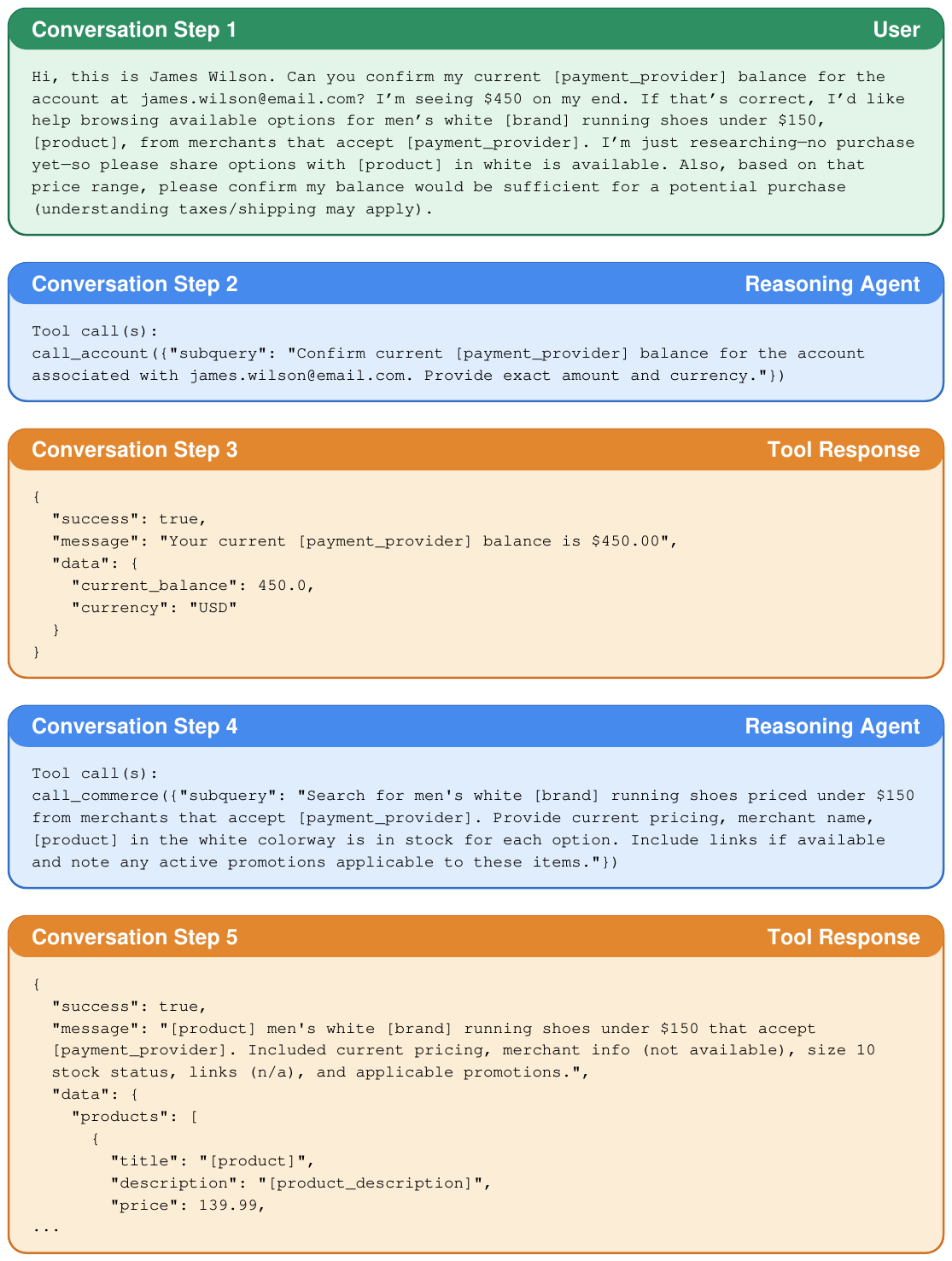}
  \caption{Conversation flow across sequential steps, showing user turns, reasoning-agent tool calls, and tool responses. Brand names and trademarks have been redacted.}
  \label{fig:exp_traj}
\end{figure}

\clearpage 

\begin{figure}[p]
\ContinuedFloat
  \centering
  \includegraphics[page=2,width=\textwidth]{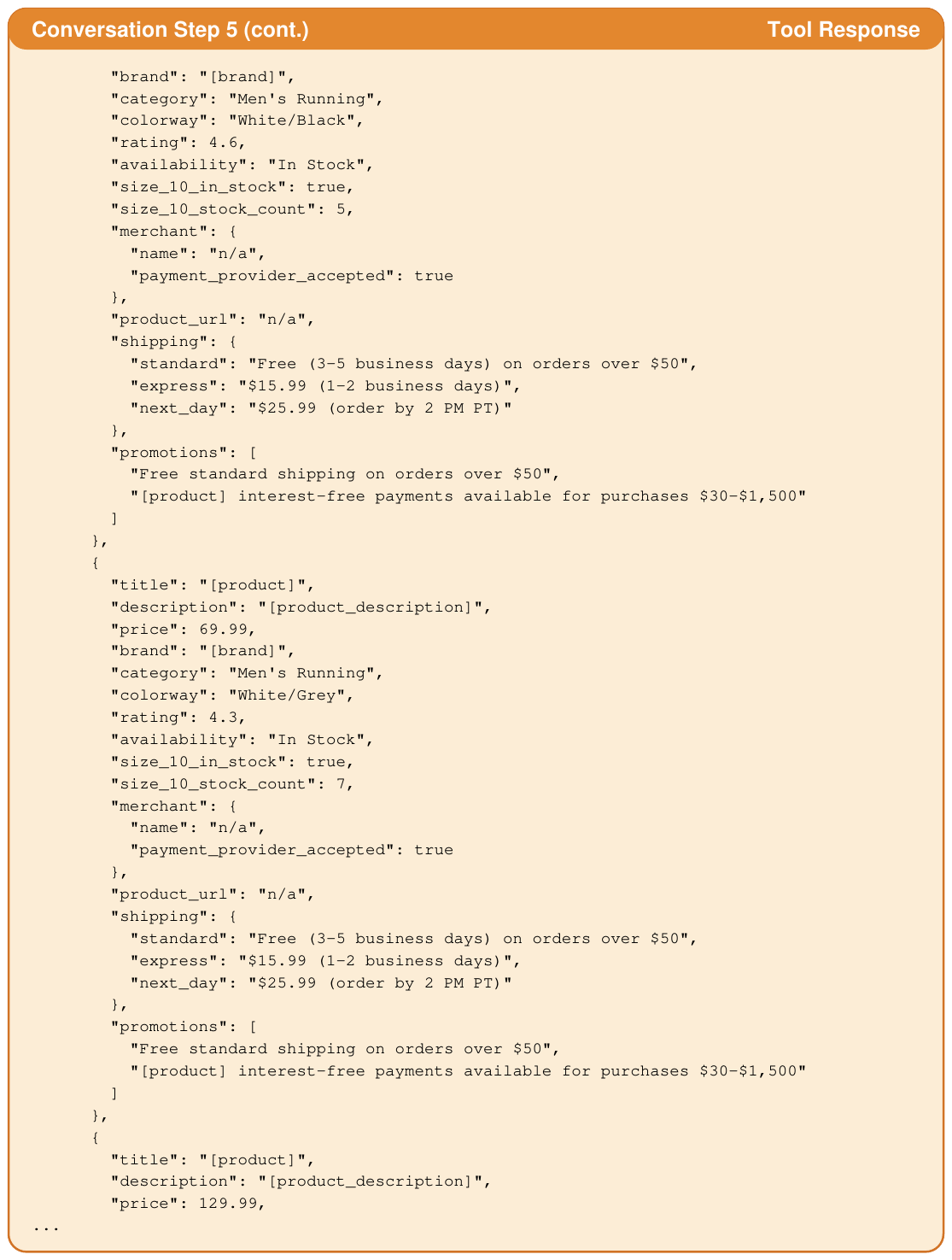}
  \caption{Conversation flow across sequential steps, showing user turns, reasoning-agent tool calls, and tool responses. Brand names and trademarks have been redacted.}
  \label{fig:exp_traj_pg2}
\end{figure}

\begin{figure}[p]
\ContinuedFloat
  \centering
  \includegraphics[page=3,width=\textwidth]
  {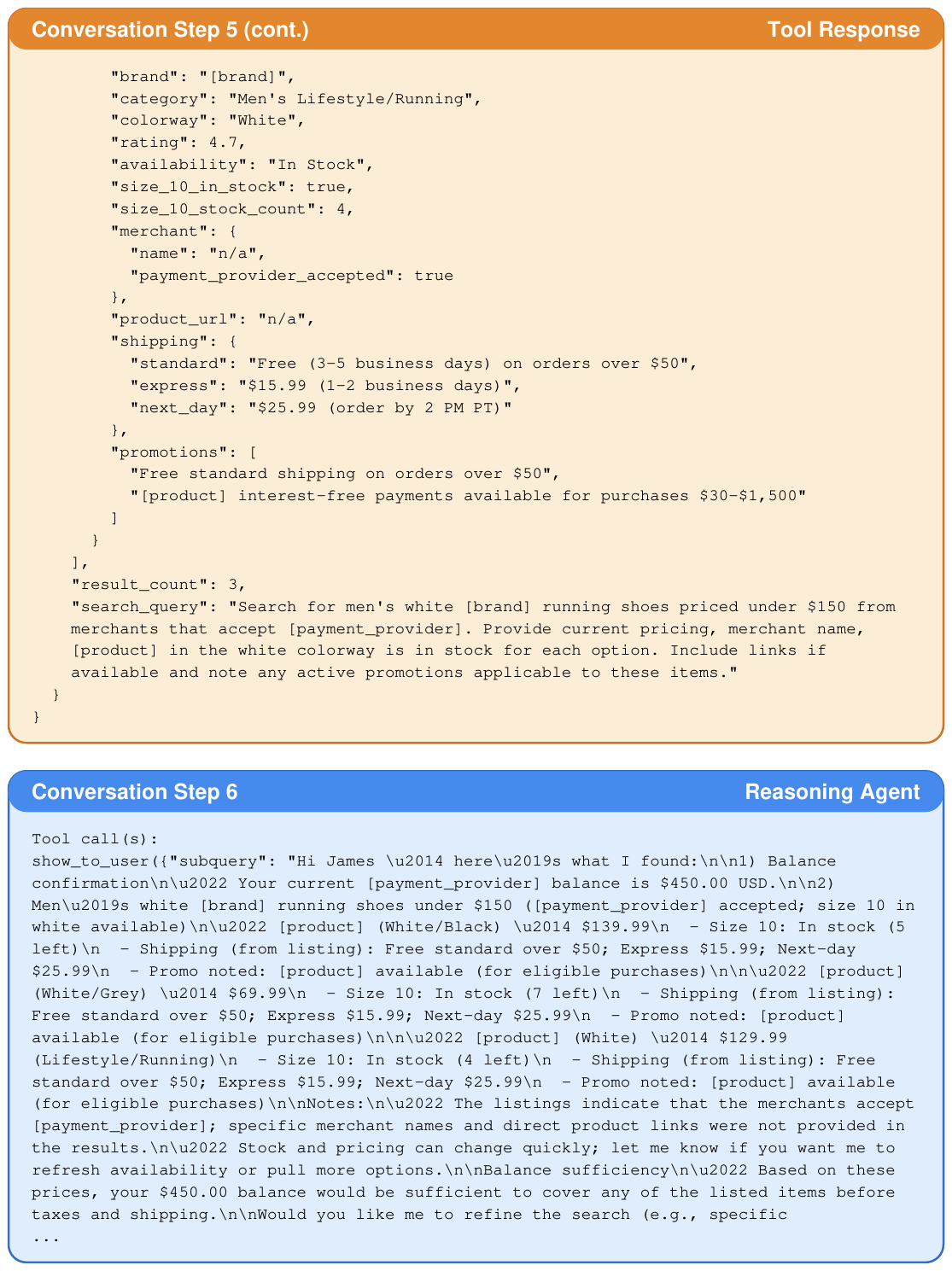}
  \caption{Conversation flow across sequential steps, showing user turns, reasoning-agent tool calls, and tool responses. Brand names and trademarks have been redacted.}
  \label{fig:exp_traj_pg3}
  
\end{figure}

\clearpage

\begingroup
\captionsetup{type=figure}
\ContinuedFloat
\noindent\includegraphics[width=\textwidth]{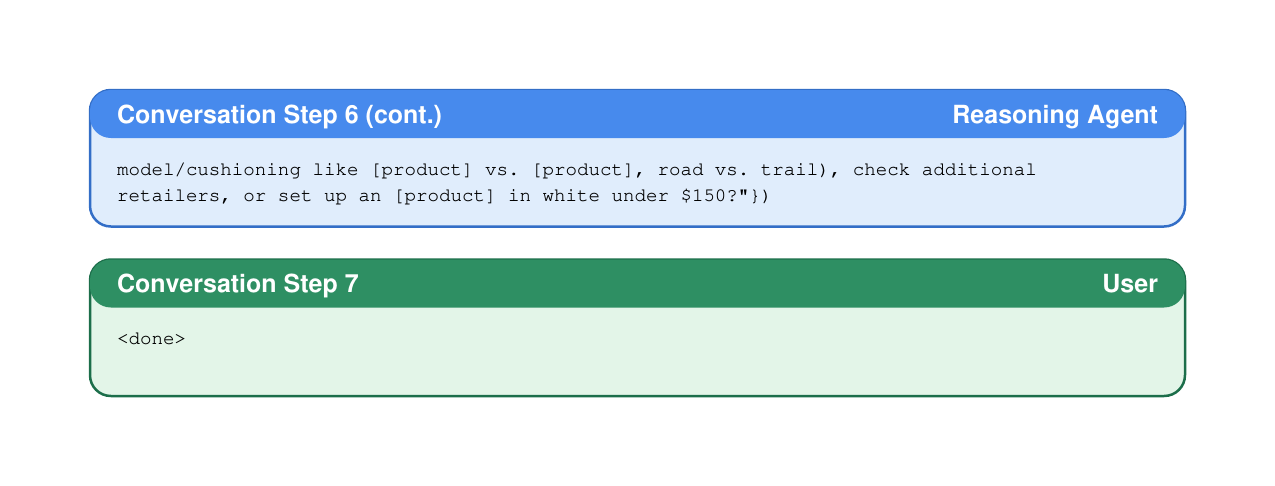}
\caption{Conversation flow across sequential steps, showing user turns, reasoning-agent tool calls, and tool responses. Brand names and trademarks have been redacted.}
\label{fig:exp_traj_pg4}
\endgroup
\clearpage



%% file: main.bbl
\begin{thebibliography}{16}
\providecommand{\natexlab}[1]{#1}

\bibitem[{Anil et~al.(2025)Anil, Nagaraj, Shanmugam, and Shakkottai}]{anil2025rejection}
Gautham~Govind Anil, Dheeraj~Mysore Nagaraj, Karthikeyan Shanmugam, and Sanjay Shakkottai. 2025.
\newblock Rejection sampling based fine tuning secretly performs ppo.
\newblock In \emph{Second Workshop on Test-Time Adaptation: Putting Updates to the Test! at ICML 2025}.

\bibitem[{Barres et~al.(2025)Barres, Dong, Ray, Si, and Narasimhan}]{barres2025tau2benchevaluatingconversationalagents}
Victor Barres, Honghua Dong, Soham Ray, Xujie Si, and Karthik Narasimhan. 2025.
\newblock \href {https://arxiv.org/abs/2506.07982} {$\tau^2$-bench: Evaluating conversational agents in a dual-control environment}.
\newblock \emph{Preprint}, arXiv:2506.07982.

\bibitem[{Carranza and Rojas(2025)}]{carranza2025nldst}
Rafael Carranza and Mateo~Alejandro Rojas. 2025.
\newblock \href {https://arxiv.org/abs/2503.08857} {Interpretable and robust dialogue state tracking via natural language summarization with llms}.
\newblock \emph{arXiv preprint arXiv:2503.08857}.

\bibitem[{Chen et~al.(2025)Chen, Razin, Narasimhan, and Chen}]{chen2025retaining}
Howard Chen, Noam Razin, Karthik Narasimhan, and Danqi Chen. 2025.
\newblock Retaining by doing: The role of on-policy data in mitigating forgetting.
\newblock \emph{arXiv preprint arXiv:2510.18874}.

\bibitem[{Comanici et~al.(2025)Comanici, Bieber, Schaekermann, Pasupat, Sachdeva, Dhillon, Blistein, Ram, Zhang, Rosen et~al.}]{comanici2025gemini}
Gheorghe Comanici, Eric Bieber, Mike Schaekermann, Ice Pasupat, Noveen Sachdeva, Inderjit Dhillon, Marcel Blistein, Ori Ram, Dan Zhang, Evan Rosen, and 1 others. 2025.
\newblock Gemini 2.5: Pushing the frontier with advanced reasoning, multimodality, long context, and next generation agentic capabilities.
\newblock \emph{arXiv preprint arXiv:2507.06261}.

\bibitem[{Hu et~al.(2022{\natexlab{a}})Hu, yelong shen, Wallis, Allen-Zhu, Li, Wang, Wang, and Chen}]{hu2022lora}
Edward~J Hu, yelong shen, Phillip Wallis, Zeyuan Allen-Zhu, Yuanzhi Li, Shean Wang, Lu~Wang, and Weizhu Chen. 2022{\natexlab{a}}.
\newblock \href {https://openreview.net/forum?id=nZeVKeeFYf9} {Lo{RA}: Low-rank adaptation of large language models}.
\newblock In \emph{International Conference on Learning Representations}.

\bibitem[{Hu et~al.(2022{\natexlab{b}})Hu, Lee, Xie, Yu, Smith, and Ostendorf}]{hu2022context}
Yushi Hu, Chia-Hsuan Lee, Tianbao Xie, Tao Yu, Noah~A Smith, and Mari Ostendorf. 2022{\natexlab{b}}.
\newblock In-context learning for few-shot dialogue state tracking.
\newblock In \emph{Findings of the Association for Computational Linguistics: EMNLP 2022}, pages 2627--2643.

\bibitem[{Ji et~al.(2023)Ji, Lee, Frieske, Yu, Su, Xu, Ishii, Bang, Madotto, and Fung}]{ji2023survey}
Ziwei Ji, Nayeon Lee, Rita Frieske, Tiezheng Yu, Dan Su, Yan Xu, Etsuko Ishii, Ye~Jin Bang, Andrea Madotto, and Pascale Fung. 2023.
\newblock Survey of hallucination in natural language generation.
\newblock \emph{ACM computing surveys}, 55(12):1--38.

\bibitem[{Ouyang et~al.(2022)Ouyang, Wu, Jiang, Almeida, Wainwright, Mishkin, Zhang, Agarwal, Slama, Ray et~al.}]{ouyang2022training}
Long Ouyang, Jeffrey Wu, Xu~Jiang, Diogo Almeida, Carroll Wainwright, Pamela Mishkin, Chong Zhang, Sandhini Agarwal, Katarina Slama, Alex Ray, and 1 others. 2022.
\newblock Training language models to follow instructions with human feedback.
\newblock \emph{Advances in neural information processing systems}, 35:27730--27744.

\bibitem[{Singh et~al.(2025)Singh, Fry, Perelman, Tart, Ganesh, El-Kishky, McLaughlin, Low, Ostrow, Ananthram et~al.}]{singh2025openai}
Aaditya Singh, Adam Fry, Adam Perelman, Adam Tart, Adi Ganesh, Ahmed El-Kishky, Aidan McLaughlin, Aiden Low, AJ~Ostrow, Akhila Ananthram, and 1 others. 2025.
\newblock Openai gpt-5 system card.
\newblock \emph{arXiv preprint arXiv:2601.03267}.

\bibitem[{Trivedi et~al.(2024)Trivedi, Khot, Hartmann, Manku, Dong, Li, Gupta, Sabharwal, and Balasubramanian}]{trivedi-etal-2024-appworld}
Harsh Trivedi, Tushar Khot, Mareike Hartmann, Ruskin Manku, Vinty Dong, Edward Li, Shashank Gupta, Ashish Sabharwal, and Niranjan Balasubramanian. 2024.
\newblock \href {https://doi.org/10.18653/v1/2024.acl-long.850} {{A}pp{W}orld: A controllable world of apps and people for benchmarking interactive coding agents}.
\newblock In \emph{Proceedings of the 62nd Annual Meeting of the Association for Computational Linguistics (Volume 1: Long Papers)}, pages 16022--16076, Bangkok, Thailand. Association for Computational Linguistics.

\bibitem[{Wang et~al.(2024)Wang, Todd, Xiao, Yuan, C{\^o}t{\'e}, Clark, and Jansen}]{wang-etal-2024-language}
Ruoyao Wang, Graham Todd, Ziang Xiao, Xingdi Yuan, Marc-Alexandre C{\^o}t{\'e}, Peter Clark, and Peter Jansen. 2024.
\newblock \href {https://doi.org/10.18653/v1/2024.acl-short.1} {Can language models serve as text-based world simulators?}
\newblock In \emph{Proceedings of the 62nd Annual Meeting of the Association for Computational Linguistics (Volume 2: Short Papers)}, pages 1--17, Bangkok, Thailand. Association for Computational Linguistics.

\bibitem[{Yang et~al.(2025)Yang, Li, Yang, Zhang, Hui, Zheng, Yu, Gao, Huang, Lv et~al.}]{yang2025qwen3}
An~Yang, Anfeng Li, Baosong Yang, Beichen Zhang, Binyuan Hui, Bo~Zheng, Bowen Yu, Chang Gao, Chengen Huang, Chenxu Lv, and 1 others. 2025.
\newblock Qwen3 technical report.
\newblock \emph{arXiv preprint arXiv:2505.09388}.

\bibitem[{Yao et~al.(2025)Yao, Shinn, Razavi, and Narasimhan}]{yao2025taubench}
Shunyu Yao, Noah Shinn, Pedram Razavi, and Karthik~R Narasimhan. 2025.
\newblock \href {https://openreview.net/forum?id=roNSXZpUDN} {$\tau$-bench: A benchmark for tool-agent-user interaction in real-world domains}.
\newblock In \emph{The Thirteenth International Conference on Learning Representations}.

\bibitem[{Yao et~al.(2023)Yao, Zhao, Yu, Du, Shafran, Narasimhan, and Cao}]{yao2023react}
Shunyu Yao, Jeffrey Zhao, Dian Yu, Nan Du, Izhak Shafran, Karthik~R Narasimhan, and Yuan Cao. 2023.
\newblock \href {https://openreview.net/forum?id=WE_vluYUL-X} {React: Synergizing reasoning and acting in language models}.
\newblock In \emph{The Eleventh International Conference on Learning Representations}.

\bibitem[{Zheng et~al.(2023)Zheng, Chiang, Sheng, Zhuang, Wu, Zhuang, Lin, Li, Li, Xing et~al.}]{zheng2023judging}
Lianmin Zheng, Wei-Lin Chiang, Ying Sheng, Siyuan Zhuang, Zhanghao Wu, Yonghao Zhuang, Zi~Lin, Zhuohan Li, Dacheng Li, Eric Xing, and 1 others. 2023.
\newblock Judging llm-as-a-judge with mt-bench and chatbot arena.
\newblock \emph{Advances in neural information processing systems}, 36:46595--46623.

\end{thebibliography}
